\documentclass[a4paper]{article}

\usepackage[top=2.8cm,bottom=3cm,left=3cm,right=3cm]{geometry}
\usepackage{graphicx}
\usepackage{amsfonts}
\usepackage{amssymb}
\usepackage{booktabs}
\usepackage{xcolor}
\usepackage{lineno}
\usepackage{textcomp}
\usepackage{amsmath}
\usepackage{natbib}

\begin{document}

%\title{Artificial Intelligence Could Have Predicted the Full Chain of Space Weather Events\\
%Associated with the May 2024 Superstorm}
%From Flare to Geomagnetic Impact}

\Large
\textbf{\begin{center}{Physics-informed features in supervised machine learning} \end{center}}

\normalsize

Margherita Lampani$^{1}$, Sabrina Guastavino$^{1,2}$, Michele Piana$^{1,2}$, and Federico Benvenuto$^{1}$ \\

\hspace{-0.5cm}$^1$ MIDA, Dipartimento di Matematica, Universit\`a di Genova, via Dodecaneso 35 16146 Genova, Italy  \\
$^2$ Istituto Nazionale di Astrofisica, Osservatorio Astrofisico di Torino, via Osservatorio 20 10025 Pino Torinese Italy \\

%\author[0000-0003-4966-8864]{Anna Maria Massone}
%\affil{University of Genoa, Department of Mathematics, Via Dodecaneso 35, I-16146 Genoa, Italy}
%\author[0000-0003-1700-991X]{Michele Piana}
%\affil{University of Genoa, Department of Mathematics, Via Dodecaneso 35, I-16146 Genoa, Italy}

\begin{center}
\textbf{Abstract}
\end{center}
Supervised machine learning involves approximating an unknown functional relationship from a limited dataset of features and corresponding labels. The classical approach to feature-based machine learning typically relies on applying linear regression to standardized features, without considering their physical meaning. This may limit model explainability, particularly in scientific applications. This study proposes a physics-informed approach to feature-based machine learning that constructs non-linear feature maps informed by physical laws and dimensional analysis. These maps enhance model interpretability and, when physical laws are unknown, allow for the identification of relevant mechanisms through feature ranking. The method aims to improve both predictive performance in regression tasks and classification skill scores by integrating domain knowledge into the learning process, while also enabling the potential discovery of new physical equations within the context of explainable machine learning.
 \\

\textbf{keywords:} supervised machine learning, feature map, physics-informed machine learning, ridge regression, feature ranking, regularization theory

\section{Introduction}
Supervised machine learning can be seen as the problem of approximating an unknown functional dependency from a finite, small set of instances made of features and the corresponding labels \citep{vapnik2013nature}. The intrinsic ill-posedness of this problem can be addressed within the framework of regularization theory \citep{kaipio2006statistical}, i.e., as the problem of minimizing a non-linear functional made of the sum of two terms: a fitting term in which the empirical risk is assessed by means of a loss function, and a penalty term that allows generalization while controlling the complexity of the solution. Finally, a real positive regularization parameter that balances the trade-off between the two terms has to be chosen by means of some regularization algorithm \citep{engl1996regularization}.

When described in a Hilbert space setting, a representer theorem \citep{scholkopf2001generalized,de2004some} provides an analytical solution of the minimum problem that
%, in the case of linear regression, 
is given by the action of a feature-dependent kernel operator onto a vector whose components can be analytically determined by means of classical Tikhonov regularization \citep{tikhonov1963solution}. 

From an operational perspective, a feature-based supervised machine learning process works as follows. Given an archive of annotated descriptors of the physical phenomenon, named features,
\begin{enumerate}
    \item A standardization procedure generates a corresponding archive of annotated standardized features that are re-scaled and made dimensionless.
    \item The machine learning algorithm is trained by leveraging the standardized features and the corresponding labels.
    \item A new set of unlabelled standardized features is fed into the trained algorithm that realizes the regression (or, after a thresholding step, the corresponding classification).
\end{enumerate}
The main limitation of this approach in scientific applications is in the fact that features represent different physical quantities, each with a distinct meaning, dimension, and unit. Although standardization techniques are computationally efficient and generally effective, they overlook the physical meanings and relationships embedded in the data, thus potentially compromising model explainability and predictive accuracy.

The objective of the present study is to introduce a novel approach to feature-based machine learning that increases its explainability by using physics-informed feature maps. In operational terms, the idea is to search for solutions of the regularization minimum problem that are linear combinations of nonlinear maps of the features, which are constructed accounting for information coming from physics, and which are triggered by the dimension of the labels. The advantage of this approach is at least two-fold. On the one hand, it fosters an a priori interpretability of the model, where the physics-informed features are homogeneous from the viewpoint of dimensions and are constructed by accounting for sound physical laws. On the other hand, when these physical laws are not known, the application of feature ranking algorithms to the physics-informed features allows the selection of the feature maps that mostly impact the prediction process, which helps to identify the possible physical mechanism at the base of data interpretation. A posteriori, one should also verify whether, in the case of regression, this approach decreases the error metrics concerned with prediction and, in the case of classification, it is able to improve the confusion matrix associated to the task, thus increasing the corresponding skill scores.

The plan of the paper is as follows. Section 2 sets up the mathematical formalism of the physics-informed approach. Section 3 describes three applications performed in synthetic settings and related to a standard regression problem, a regression problem where the physics-informed feature corresponding to the corrected physical equation is removed, and a standard classification problem. Section 4 focuses on an experiment leveraging real data concerned with a space weather problem. Our conclusions are offered in Section 5.

\section{Physics-informed feature maps}

The aim of supervised learning is to find a function $g: \Omega \subset \mathbb R^m \longrightarrow \mathbb R$ from a set of examples $\{ (x_i,y_i) \}_{i=1,\ldots,n}$ randomly drawn from an unknown probability distribution $\rho(x,y)$ with $( y_1,\ldots , y_n ) \in \mathbb R^n$ and $( x_1,\ldots x_n ) \in \Omega^n$, such that $g$ has to explain the relationship between input and output, i.e.
\begin{equation}\label{Y sim g(X)}
    y_i \sim g(x_i)
\end{equation}
for all $i=1,\dots,n$ and $g(x)$ has to be a good estimate of the output when a new input $x \in \Omega$ is given. A classical setting is given by regression, where it is usually assumed that 
\begin{equation}
    y_i=g(x_i)+\epsilon_i
\end{equation}
where $\epsilon_i$ is Gaussian noise, and the sought solution $g$ is a linear function of the form $g(x)=x^T w$, with $w\in\mathbb{R}^m$ being the vector of coefficients to estimate and $x \in \Omega$. 
This problem is typically studied using variational approaches \citep{zhang2018advances}, that is by minimizing the empirical expected value of the loss function over $w\in\mathbb{R}^m$. 
Since the minimizing solution can be numerically unstable \citep{poggio1984ill}, it is common to apply Tikhonov regularization \citep{golub1999tikhonov}, introducing a penalty term to stabilize the solution. 
The regularized counterpart of the problem is then formulated as the minimization problem
\begin{equation}
    \hat{w} =\arg\min_{w\in\mathbb{R}^m} \sum_{i=1}^n V(y_i,x_i^T w) + \lambda\Vert w\Vert_2^2,
\end{equation}
where $V$ is the loss function \citep{rosasco2004loss} expressesing a distance between the noisy data $y_i$ and their estimated values for each $i$, and $\lambda>0$ is the regularization parameter. Using the classical squared loss function, this leads to ridge regression (also known as penalized least squares) \citep{mcdonald2009ridge}, whose explicit solution is given by
\begin{equation}
    \hat{w} = (\textbf{X}^T\textbf{X} + \lambda I)^{-1} \textbf{X}^T \textbf{Y},
\end{equation}
where $\mathbf{X}$ represents the $n\times m$ design matrix with $x_i$ as rows, $\textbf{Y}$ is the $n\times 1$ vector whose components are the features' lables, and $I$ denotes the identity matrix . 

However, the relationship between input features and regression output may be more complex than the one modeled by a linear scenario. In such cases, feature maps and kernels provide a way to generalize linear models to non-linear frameworks. In fact, a feature map $\phi:\Omega\to\mathcal{F}$ maps the input space $\Omega$ of the original features to a new feature space $\mathcal{F}$ where a scalar product is defined. This transformation allows the data to be represented in a space where separation performed by means of a linear model is easier. 
For example, if $\mathcal{F}= \mathbb{R}^p$, the goal is to find a function $g(x_i)=\phi(x_i)^T\beta$, where $\beta\in \mathbb{R}^p$ is the ideal coefficient vector. From a practical perspective, the problem is addressed by solving a similar variational problem, where the ideal coefficient $\beta$ is estimated as
\begin{equation}\label{eq: hat beta}
    \hat{\beta} = (\Phi^T\Phi + \lambda I)^{-1} \Phi^T \mathbf{Y},
\end{equation}
where $\Phi$ is the feature map matrix $n\times p$ whose rows are given by a set of feature maps $\{\phi(x_i)\}$. This approach is closely related to kernel methods, since a kernel can be defined through a feature map, i.e. by defining $k(x_i, x_l)=\langle \phi(x_i), \phi(x_l)\rangle_{\mathcal{F}}$.

In the kernel framework, a kernel function (such as a Gaussian or polynomial function) is typically chosen without accounting for the physical meaning or dimensional aspects of the features $x_i$. 
%This limitation can be addressed by constructing a new feature map $\phi$ which combines the original features according to physics-driven laws, resulting in dimensionally homogeneous, physics-informed features.
%In standard approaches, the feature maps are chosen without accounting for the physical meaning or dimensional properties of the features $\{x_i\}$. 
Our goal was to address this limitation by building new feature maps $\phi$ that combine the original features according to physics-driven laws, resulting in dimensionally homogeneous, physics-informed features (PIFs).
Just as an example, let us consider $x=(m, v, E)$, where $m$ represents a mass, $v$ a speed, and $E$ an energy. 
We construct the following physics-informed feature map $\phi:\mathbb{R}^3\to\mathbb{R}^3$
\begin{equation}
    \phi(x) = (m v^2, E, m^2 v^4 /E),
\end{equation}
and therefore, the forward model is represented by
\begin{equation}
   g(x) = \beta_1 m v^2 + \beta_2 E + \beta_3 m^2 v^4 /E.
\end{equation}
Machine learning here has therefore the two-fold task to compute the right values for $\beta_1$, $\beta_2$ and $\beta_3$, and to select the most predictive PIFs by means of some feature ranking algorithm. 

We would like to remark now that, thanks to the Reproducing Kernel Hilbert Space (RKHS) theory \citep{berlinet2011reproducing},  finding the function $g$ that links input and output is equivalent to solving a linear inverse problem where the forward operator $A$ is defined by the physics-informed feature map. In the following section, we provide this formalization in a more general setting where general Hilbert spaces are considered.

\subsection{Connection between physics-informed forward operator and Reproducing Kernel Hilbert Spaces}
\label{sec:kernel}
In machine learning or, more in general, in approximation theory, the function $g$ linking the input features to the output labels is usually assumed to belong to an RKHS, where a reproducing kernel is fixed.
In this more general setting, this problem is typically addressed by solving a variational problem in which the loss function $V$ is arbitrary, and the regularization term generalizes the classical Tikhonov penalty term, depending on the norm $\| \cdot \|_{\mathcal{H}_K}$ in the RKHS $\mathcal{H}_K$. Therefore the following variational problem is considered:
\begin{equation}
\label{approx_probl_sampling_noisy}
\hat{g}_{\lambda}:=\arg\min_{g\in\mathcal{H}_K} \sum_{i=1}^n V(y_i, g(x_i))+\lambda\psi(\| g\|_{\mathcal{H}_K})~,
\end{equation}
where $\lambda$ is the regularization parameter and $\psi:[0,+\infty)\to\mathbb{R}_+$ is a continuous convex and strictly monotonically increasing real-valued function.

A well-established result proves that an RKHS can be viewed as the image space of a compact linear operator $A$ (e.g., see \citep{steinwart_support_2008,kress1989linear, aronszajn1950theory}).
%, to which we can associate an inverse problem. 
The following proposition is an adaptation of this result in the context of feature maps.

\noindent
{\bf{Proposition 2.1}}
{\em{Let us consider a compact linear operator $A:{\mathcal{H}}_1 \to {\mathcal{H}}$ where $\mathcal{H}_1$ is a Hilbert space and $\mathcal{H}$ is a Hilbert space of functions over $\Omega$.
In this case, for all $x$ there exists an element $\phi(x)\in\mathcal{H}_1$ such that
\begin{equation}
\label{scalar-product}
(Af)(x) = \langle  f, \phi(x)\rangle_{\mathcal{H}_1},
\end{equation}
for each element $f \in \mathcal H_1$.
Then the range of the operator $A$, denoted as
$\Im(A)$, equipped with the norm
\begin{equation}
\| g\|_{\mathcal{H}_K} = \min\{ \| w\|_{\mathcal{H}_1} : w\in\mathcal{H}_1 \ \ s.t. \ \  g(x)=\textlangle w,\phi(x)\textrangle_{\mathcal{H}_1},  ~ x\in \Omega\}
\end{equation}
is an RKHS 
with kernel
\begin{equation}\label{Kernel}
\begin{array}{c}
K :   \Omega\times \Omega\to \mathbb{R} \\ \\
  (x,r) \to K(x,r):= \langle \phi(x),\phi(r)\rangle_{\mathcal{H}_1}.
  \end{array}
\end{equation}
}}
We remark that
\begin{equation}
\Im(A) = \overline{span\{ K(x,\cdot), ~ x\in\Omega\}} ~ ,
\end{equation}
and that the feature space 
\begin{equation}
\mathcal{F}:=\overline{span\{\phi(x) ~, ~ x\in \Omega\}}
\end{equation}
is such that $\mathcal F = \ker(A)^{\perp}$.
In summary, the first isomorphism theorem applied to the map
\begin{equation}
A : \mathcal H_1 \longrightarrow \Im(A) = \mathcal H_K \subseteq \mathcal H
\end{equation}
states that
\begin{equation}\label{eq:summary}
\mathcal F = \ker(A)^{\perp} \simeq {\mathcal H_1}/{\ker(A)} \simeq \Im(A) = \mathcal H_K ~.
\end{equation}

Proposition 2.1 and equation (\ref{eq:summary}) allowed us to provide an equivalence result between the approximate solution (\ref{approx_probl_sampling_noisy}) and the regularized solution
\begin{equation}
\label{inv_probl_sampling_noisy}
\hat{f}_{\lambda}:= \arg\min_{f\in\mathcal{H}_1} \sum_{i=1}^n V(y_i, (Af)(x_i))+\lambda \psi(\| f\|_{\mathcal{H}_1})
\end{equation}
of the discretized inverse problem $Af = y$, where $A$ is the operator characterized by the map $\phi$ defined in (\ref{scalar-product}), and $y_i$ for $i=1,\ldots,n$, are noisy samples of the ideal data $y$. Indeed, the following result holds true:

\noindent
{\bf{Proposition 2.2.}} {\em{Given $\{(x_i, y_i)\}_{i=1}^n$ a set of samples and  by assuming that $K(x,x')=\langle \phi(x),\phi(x')\rangle_{\mathcal{H}_1}$ for all $x,x'\in\Omega$ we have that
\begin{equation}\label{correspondence discretization}
\hat{f}_{\lambda}
= A^{\dagger} 
\hat{g}_{\lambda},
\end{equation}
where $A^{\dagger}: \Im(A)+\Im(A)^{\perp} \to \ker(A)^{\perp}$ is the generalized inverse of $A$.}}\\
{\bf{Proof.}}
By hypothesis we have the identification $\Im(A)=\mathcal{H}_K$. 
Let $\tilde{f} := A^{\dagger} \hat{g}_{\lambda}$.
By definition of $A^{\dagger}$, $\tilde{f} \in \ker(A)^{\perp}$ and therefore, $\| \hat{g}_{\lambda}\|_{\mathcal{H}_K}=\| \tilde{f} \|_{\mathcal{H}_1}$. 
For all $f \in \mathcal{H}_1$ we have
\begin{equation}
\begin{array}{c}
\sum_{i=1}^n V(y_i,(Af)(x_i)) +\lambda \psi(\| f\|_{\mathcal{H}_1})  \ge \min_{g\in \Im(A)} \sum_{i=1}^n V(y_i,g(x_i)) +\lambda \psi(\| g\|_{\mathcal{H}_k})\\ \\
=  \sum_{i=1}^n V(y_i,\hat{g}_{\lambda}(x_i))+\lambda \psi(\| \hat{g}_{\lambda} \|_{\mathcal{H}_k})
= \sum_{i=1}^n V(y_i,A \tilde{f}(x_i))+\lambda \psi(\| \tilde{f} \|_{\mathcal{H}_1}) \ \ ,
\end{array}
\end{equation}
i.e. $\tilde{f}$ is solution of problem (\ref{inv_probl_sampling_noisy}). This concludes the proof.

\begin{figure}[htbp]
  \centering
  \includegraphics[width=0.9\textwidth]{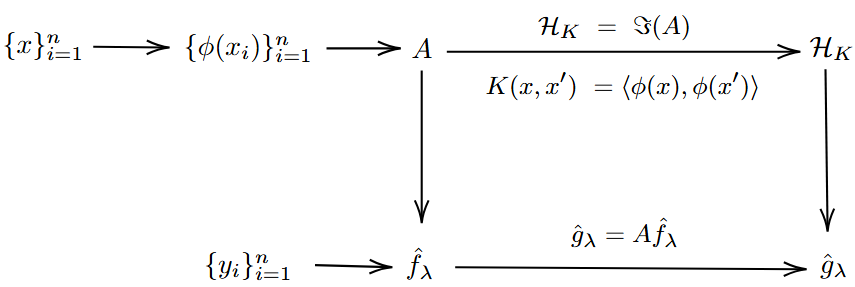}
  \caption{Outline of the process that implements the physics-driven solution of the feature-based machine learning problem: the measured features are transformed by a physics-informed map into an operator $A$ in a Hilbert space setting; the solution $\hat{f}_{\lambda}$ of the corresponding inverse problem provides the physical equation, which is transformed by $A$ into the predicted output $\hat{g}_{\lambda}$.} %Scheme illustrating the equivalence between the machine learning (or
%approximation) problem in an RKHS, and the linear inverse problem through the connection between the physics-informed feature map $\phi$ and the reproducing kernel $K$.}
  \label{fig:scheme}
\end{figure}

The connection between $\hat{f}_{\lambda}$ and $\hat{g}_{\lambda}$ can be leveraged and interpreted within the context of the proposed physics-informed approach as in Figure \ref{fig:scheme}. This figure explains that, given a set of samples, we construct a physics-informed feature map by using some knowledge of the feature input (such as the physical meaning or dimensional properties). This can be viewed as defining a linear operator $A$, which encodes the physical model equation, being defined via the physics-informed feature map $\phi$ (see equation (\ref{scalar-product})). Solving the inverse problem associated with this operator is equivalent to solving a machine learning or approximation problem, where the solution is in a specific Reproducing Kernel Hilbert Space (RKHS) $\mathcal{H}_K$, with the kernel $K$ defined by the physics-informed feature map $\phi$, i.e., $
K(x, x') = \langle \phi(x), \phi(x') \rangle_{\mathcal{H}_1}
$
for all $x, x' \in \mathcal{X}$.

%\begin{figure}

%\caption{Commutative diagram summarizing the equivalence between approximation in a RKHS and linear inverse problems.}\label{equivalence-diagram}
%\end{figure}

%Two remarks on this result, reinterpreted in the context of the proposed physics-informed approach, are needed.

\section{Experiments in synthetic settings}\label{sec2}

The basic idea of this section is to show how physics-informed features can be utilized to construct physically sound models. We considered three different theoretical frameworks, i.e. three different model equations and, for each one, 
\begin{enumerate}
    \item We generated an annotated archive of features, where the corresponding labels are computed by solving the model equation and by applying uniform random noise of different intensities.  
    \item We split the annotated archive into a training set and a test set.
    \item We applied standardization to generate the corresponding training and test sets of standardized features (SFs).
    \item We trained a machine learning algorithm and then applied it to the test set, accordingly computing evaluation metrics such as the mean absolute error (MAE) and the mean squared error (MSE).
\end{enumerate}
Then, we used the same archive of features to generate a training and a test set of PIFs by means of dimensionally homogeneous combinations of the original features, where the dimension of reference is the one of the corresponding labels (see Algorithm 3.1); the PIFs were then standardized to generate the corresponding set of standardized physics-informed features (SPIFs); the machine learning algorithm optimized by means of the training set of SPIFs was validated by using the corresponding test set; and, finally, evaluation metrics have been computed for comparison with the results obtained using SFs in the training phase. In all experiments considered below, the annotated archives are made of $1000$ points in both the SFs and PIFs spaces and, in all cases, $70\%$ of them were used for training and $30\%$ for testing. Further, we used ridge regression as the machine learning algorithm, although we found that the results are robust while using other machine learning approaches (specifically, the results obtained by means of a support vector machine algorithm for regression were very close to the ones obtained with ridge regression).

In order to provide PIFs with a ranking that describes their impact on regression, we applied a greedy approach to feature ranking \citep{camattari2024classifier} (see Algorithm 3.2), where
\begin{enumerate}
    \item SPIFs are ranked according to the values of the computed regression coefficients.
    \item Recursively and starting from the SPIF with highest rank, the prediction is performed up to saturation of MAE and MSE.
    \item The regression coefficients of the selected SPIFs are then de-standardized in order to determine the (physically meaningful) regression coefficients associated to the corresponding PIFs.
\end{enumerate}
This last step allows the most significant physical equation used to generate the PIFs to be identified.

\noindent
{\bf{Remark 3.1.}} We point out that the standardization/de-standardization process of the PIFs reduces possible numerical issues in the computation of the weights while allowing the determination of the physical coefficients associated to the model equations.

\begin{figure}
\includegraphics[width=1\textwidth]{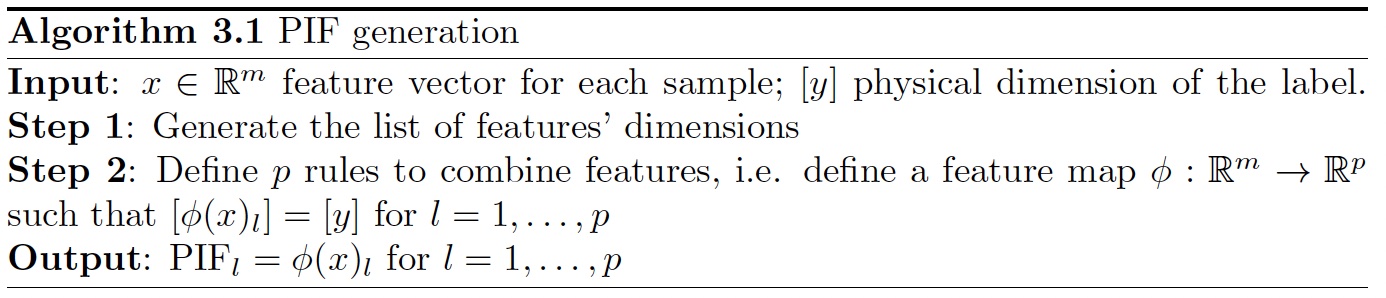}
\end{figure}

\begin{figure}
\includegraphics[width=1\textwidth]{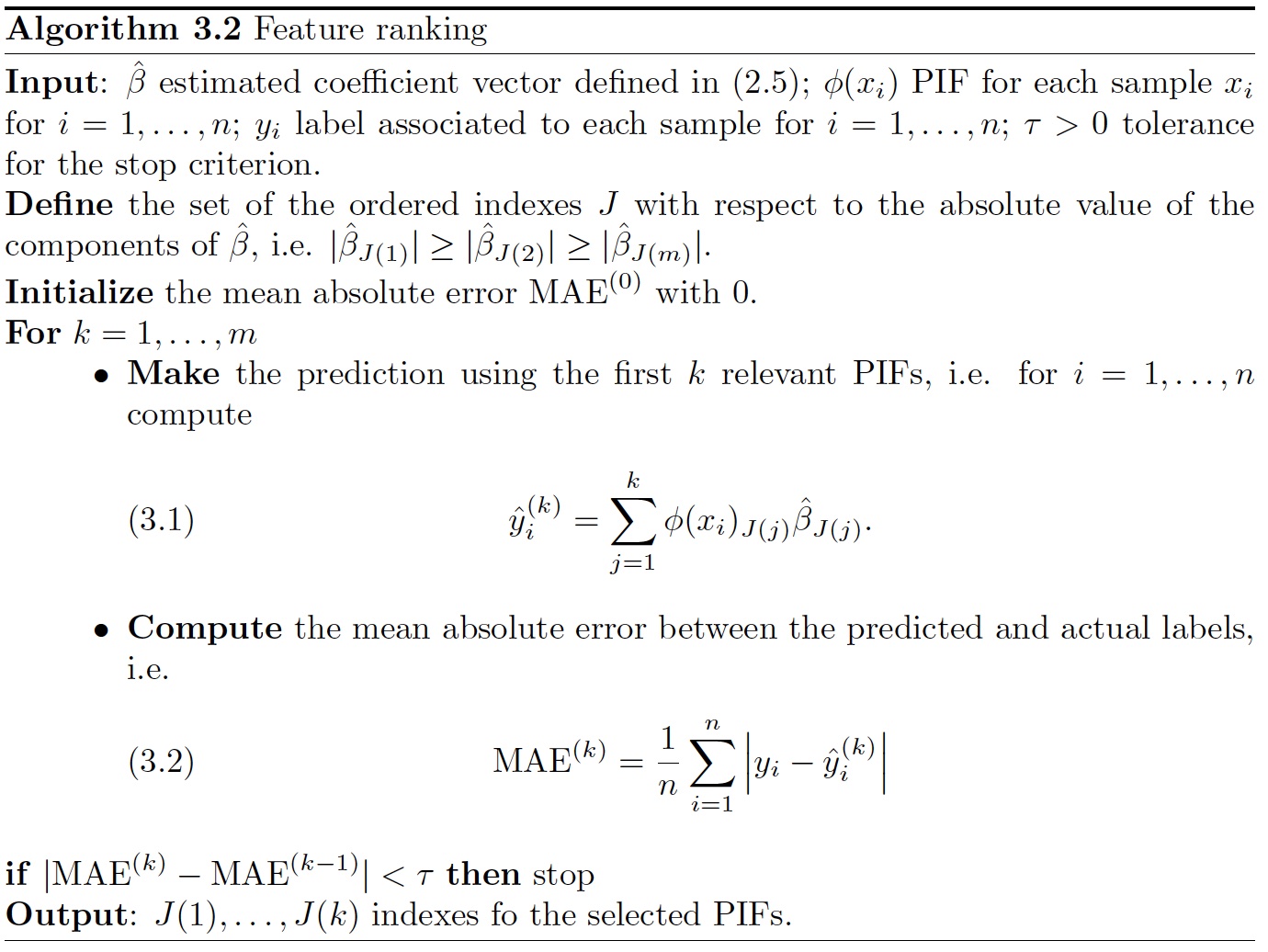}
\end{figure}

\subsection{First example: fluid dynamics}
Bernoulli equation \citep{anderson1995computational} represents the fundamental law that models the dynamics of the ideal fluid under stationary flow conditions. It is expressed as
\begin{equation}\label{eq:bernoulli}
    p + \frac{\rho v^2}{2} + \rho g h = const,
\end{equation}
where $\rho$ is the fluid density, $v$ is its speed, $p$ denotes the static pressure of the fluid (either water or air in this experiment), and $\frac{\rho v^2}{2}$ is the dynamic pressure arising from the fluid's motion; finally, $\rho g h$ corresponds to the hydrostatic pressure exerted by a column of fluid, where $g$ is the gravitational acceleration and $h$ is the height of the fluid column. Together with the aforementioned features, we incorporated additional features essential for the study of a fluid dynamic system, i.e., the volumetric flow rate $Q$, the cross-sectional area of the column $A$, and the dynamic viscosity $\mu$. We combined these seven features to generate seven PIFs characterized by the dimension of a pressure, where the first three PIFs are the ones incorporated in the Bernoulli equation (\ref{eq:bernoulli}). From a mathematical view point, in this example $x = (p,\rho, v, Q, A, \mu, h)$ and we defined the physics-informed feature map $\phi: \mathbb{R}^7 \to \mathbb{R}^7$ such as 
\begin{equation}
    \phi(x) = (p, \rho v^2, \rho g h, \frac{pQ}{Av}, \frac{\mu v}{h}, \frac{\mu Q}{h A^2}, \frac{\rho A g}{h}).
\end{equation}
Table \ref{tab:bernoulli-features} contains the definitions of all original features and PIFs together with the corresponding units.

\begin{table}[h]
    \centering
    \begin{tabular}{ l c l c}
    \toprule
      {\bf{feature}} & {\bf{unit}}  & {\bf{PIF}}                 & {\bf{unit}} \\ \midrule
      F$_1=p$    & $\frac{kg}{m \cdot s^2}$ & PIF$_1 = p$                    & Pa \\ \midrule
      F$_2=\rho$ & $\frac{kg}{m^3}$         & PIF$_2 = \rho v^2$             & Pa \\ \midrule
      F$_3=v$    & $\frac{m}{s}$            & PIF$_3 = \rho g h$             & Pa \\ \midrule
      F$_4=Q$    & $\frac{m^3}{s}$          & PIF$_4 = \frac{pQ}{Av}$        & Pa \\ \midrule
      F$_5=A$    & $m^2$                    & PIF$_5 = \frac{\mu v}{h}$      & Pa \\ \midrule
      F$_6=\mu$  & $\frac{kg}{m \cdot s}$   & PIF$_6 = \frac{\mu Q}{h A^2}$  & Pa \\ \midrule
      F$_7=h$    & $m$                      & PIF$_7 = \frac{\rho A g}{h}$   & Pa \\ \bottomrule
    \end{tabular}
    \caption{Prediction of Bernoulli equation. Features and PIFs generated by a feature map triggered by the pressure dimension. PIF$_1$, PIF$_2$, and PIF$_3$ are the PIFs that generate equation (\ref{eq:bernoulli}).}
    \label{tab:bernoulli-features}
\end{table}

\begin{figure}
\includegraphics[width=1\textwidth]{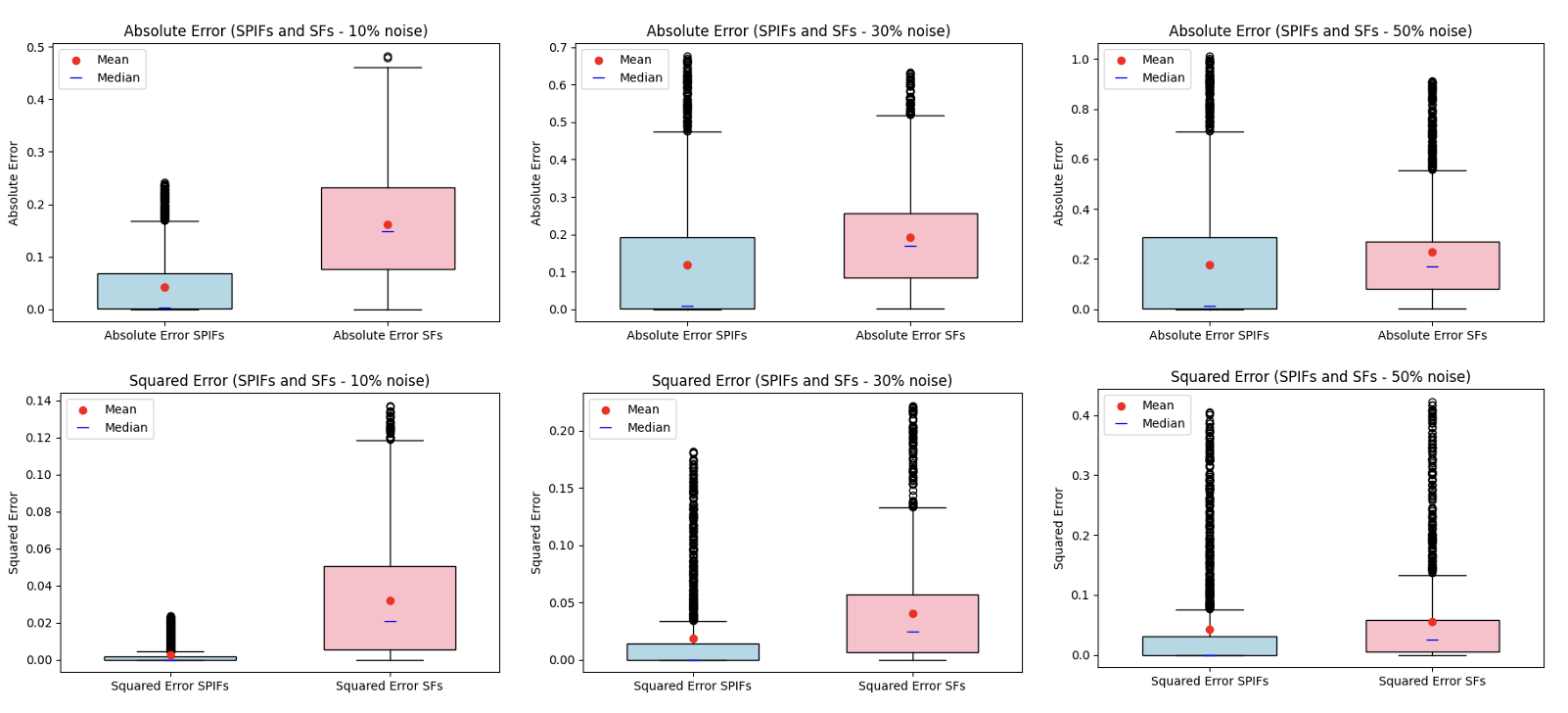}
\caption{Boxplots of the IQR distributions for absolute errors (top panel) and squared errors (bottom panel) provided by ridge regression for predicting the Bernoulli equation, by varying the noise levels. The light blue and pink boxlots represent the results obtained using SPIFs and SFs, respectively.}\label{fig:bernoulli-metrics}
\end{figure}

The results of the application of ridge regression using both SFs and SPIFs are given in Table \ref{tab:bernoulli-metrics} and Figure \ref{fig:bernoulli-metrics}. Table \ref{tab:bernoulli-metrics} contains the Mean Absolute Errors (MAEs) and the Mean Squared Errors (MSEs) provided by ridge regression when using SFs and SPIFs as features. Figure \ref{fig:bernoulli-metrics} represents the corresponding box-plots where, for a better visualization, we included just the inter-quartile range (IQR box-plots from now on).

\begin{figure}
    \centering   \includegraphics[width=1\linewidth]{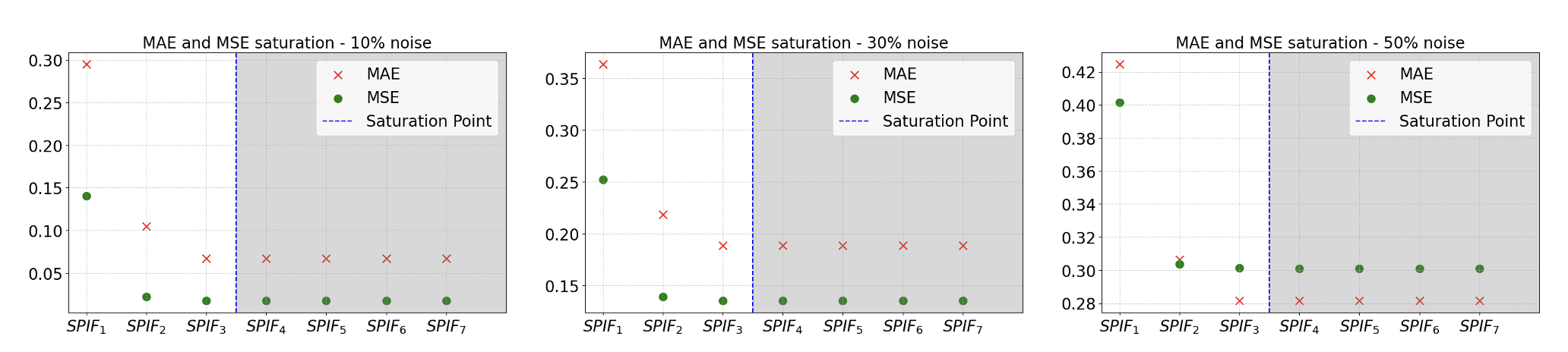}
    \caption{Prediction of the Bernoulli equation. Saturation of MAE and MSE while applying the sequential feature ranking Algorithm 3.2.}
    \label{fig:bernoulli-saturation}
\end{figure}

We then applied the feature ranking algorithm to SPIFs and showed in Figure \ref{fig:bernoulli-saturation} that MAE and MSE saturation occurs starting from SPIF$_4$. The de-standardization of the coefficients associated to SPIF$_1$, SPIF$_2$, and SPIF$_3$ leads to the physics-informed regression coefficients in Table \ref{tab:bernoulli-ranking}. These coefficients are clearly an accurate approximation of the physical coefficients in the Bernoulli equation (\ref{eq:bernoulli}).

%which contains the box-plots for Absolute Errors and Squared Errors. Further, Table \ref{tab:fluid-dynamic-rank} shows that the PIFs selected by machine learning with the highest rank are $PIF_1$, $PIF_2$, and $PIF_3$, i.e. the $SF$s included in Bernoulli equation, and that the corresponding regression coefficients are very close to the physical coefficients in equation (\ref{eq:bernoulli}).

\begin{table}[]
\centering
\begin{tabular}{l  c|c|c|c|c|c}
\toprule
            & \multicolumn{6}{c}{noise level} \\  \cmidrule(lr){2-7}
            & \multicolumn{2}{c|}{10\%} & \multicolumn{2}{c|}{30\%} & \multicolumn{2}{c}{50\%} \\ \midrule
metric      & SFs             & SPIFs         & SFs             & SPIFs        & SFs             & SPIFs          \\ \midrule
MAE         & 0.179           & 0.067         & 0.275           & 0.189        & 0.353           & 0.281          \\ \midrule
MSE         & 0.052           & 0.017         & 0.168           & 0.135        & 0.330           & 0.301          \\ \bottomrule
\end{tabular}
\caption{MAE and MSE computed from the results provided by ridge regression when SFs and SPIFs are used to predict the Bernoulli equation (\ref{eq:bernoulli}). Three different levels of uniform noise were applied to the synthetic labels.}
\label{tab:bernoulli-metrics}
\end{table}

%Examining Table \ref{tab:bern metrics_results}, we observe that the operations once again outperform the features. Furthermore, the MAE saturation occurs when the three key components of the equation ($Op_1$, $Op_2$, $Op_3$; see Table \ref{tab: bern coefficients}) are included, demonstrating that the most relevant operations are identified. Moreover, their coefficients are accurately identified and extracted (Table \ref{tab: bern coefficients}).

\begin{table}[h!]
\centering
\begin{tabular}{c  c | c | c  c}
\toprule
& \multicolumn{3}{c}{noise level} \\ \cmidrule(lr){2-4}
& 10\% & 30\% & 50\% & \\ \midrule
& \multicolumn{3}{|c|}{estimated coefficient} & \multicolumn{1}{|c}{ground-truth coefficient} \\ \midrule
SPIF$_1$   &  0.998   & 0.993   & 0.990  & 1   \\ \midrule
SPIF$_2$   &  0.500   & 0.501   & 0.502  & 0.5 \\ \midrule
SPIF$_3$   &  1.005   & 1.024   & 1.031  & 1   \\ \bottomrule
%$PIF_4$  &  0.0002 & 0.0007 & 0.001 & - \\ \hline
%$PIF_5$ & 22070.045 & 59537.509 & 105779.681  & - \\ \hline
%$PIF_6$ & -19937.350  & -57521.637 &  -98015.686 & - \\ \hline
%$PIF_7$ & 0.403  & 1.266 &2.053 & - \\ \hline
\end{tabular}
\caption{Prediction of the Bernoulli equation. Best ranked PIFs and corresponding regression coefficients for three levels of uniform noise affecting the labels.}
\label{tab:bernoulli-ranking}
\end{table}

%\begin{figure}[]
%\includegraphics[width=1\textwidth]{BERNOULLI SENZA IQR.png}
%\caption{---SENZA USARE IQR}\label{fig:fluid-dynamic}
%\end{figure}

\subsection{Second example: magnetic dissipation in energetic pulsars}

Pulsars are highly magnetized rotating neutron stars that emit beams of electromagnetic radiation \citep{weber2017pulsars}. Through these emissions, they loose magnetic energy according to the dissipation law \citep{hakobyan2023magnetic}
\begin{equation} \frac{dE}{dt} = -\frac{2\pi B^2 r^6 \omega^4 \sin^2 \alpha}{3\mu_0 c^3}, 
\label{eq:pulsar}
\end{equation}
 where $B$ is the magnetic field of the star, $r$ is its radius, $\omega$ is its angular velocity, $\alpha$ is its inclination angle, $\mu_0$ is the magnetic permeability in vacuum and $c$ is the speed of light in vacuum. In addition to these quantities, we introduced four more representative features of the system: $P = \frac{2\pi}{\omega}$, the rotational period of the star; $m$, its mass; $I = \frac{2mr^2}{5}$, the moment of inertia; and $E = \frac{I\omega^2}{2}$, the rotational kinetic energy.
 As usual, PIFs were defined according to the physical unit of the label, i.e., power, and, again, PIF$_1$
 represents the exact equation for the magnetic energy dissipation as in (\ref{eq:pulsar}) but excluding the dimensionless coefficients. Both features and PIFs are summarized in Table \ref{tab:pulsar}.

\begin{table}[h]
    \centering
    \begin{tabular}{ l c l c}
    \toprule
      \bf{feature} & \bf{unit}   & \bf{PIF} & \bf{unit} \\ \midrule
      F$_1=r$ & $m$ & PIF$^{\ast}_1 =  -\frac{B^2r^6\omega^4 sin^2(\alpha)}{\mu_0c^3}$ & W  \\ \midrule
      F$_2=B$ & $T$ & PIF$_2 = -\frac{B^2c^4 sin^2(\alpha)}{\omega^3\mu_0} v^2$ & W \\ \midrule
      F$_3=\omega$ & $\frac{1}{s}$ & PIF$_3 = -\frac{B^2r^4\omega}{\mu_0}$ & W \\ \midrule
      F$_4=\alpha$ & rad & PIF$_4 =  -E/P$ & W \\ \midrule 
      F$_5=P$ & $S$ & PIF$_5 = -I/P^3$ & W \\ \midrule
      F$_6=m$ & $kg$ & PIF$_6 = - mr^2\omega^3$ & W \\ \midrule
      F$_7=I$ & $kg\cdot m^2$ & PIF$_7 = -\frac{mr^2}{\omega^3}$ & W \\ \midrule
      F$_8=E$ & $\frac{kg\cdot m^2}{s^2}$ & & \\ \bottomrule
    \end{tabular}
    \caption{Prediction of the pulsar equation (\ref{eq:pulsar}). Features and PIFs with respective units. The $\ast$ for PIF$_1$, which corresponds to the correct physical equation, denotes the fact that the second run of the experiment has been carried out without that PIF.}
    \label{tab:pulsar}
\end{table}

\begin{figure}    \includegraphics[width=1\textwidth]{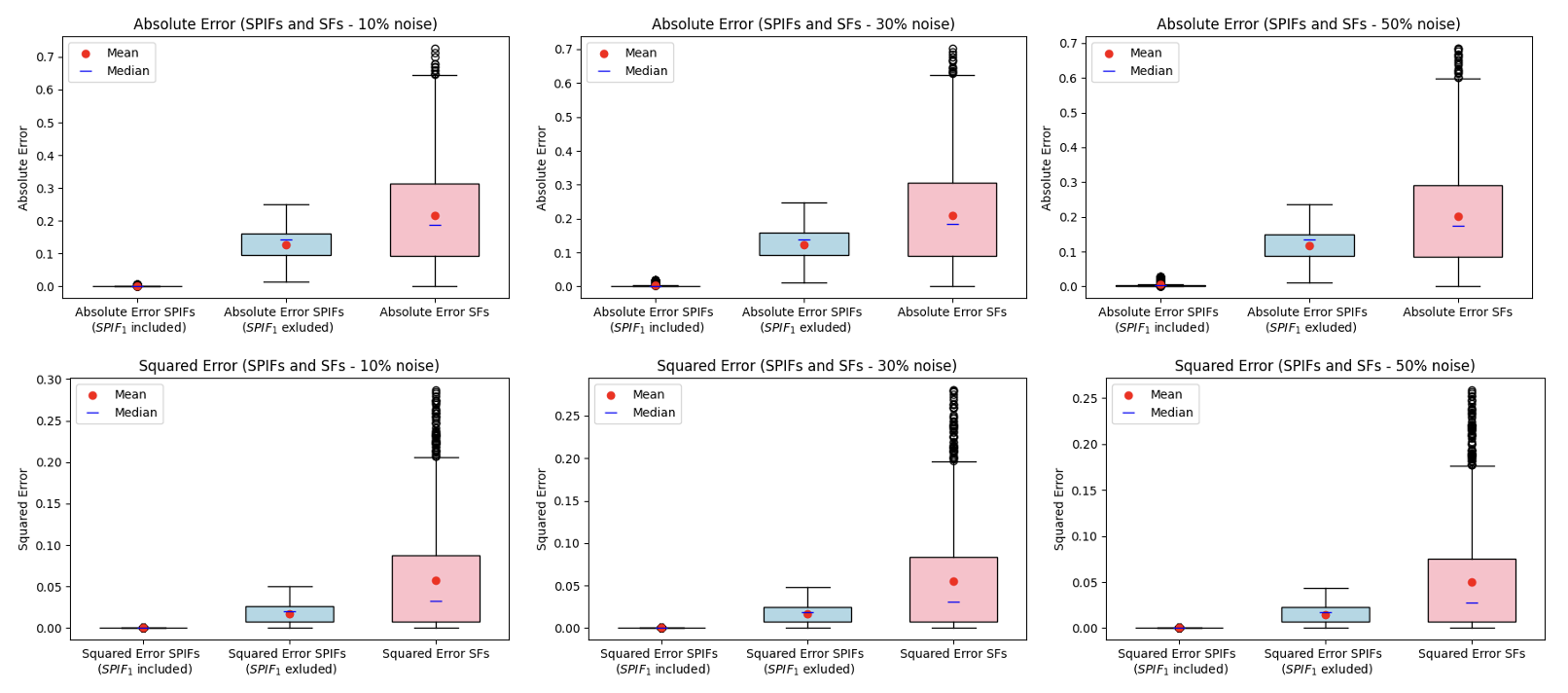}
    \caption{Prediction of the pulsar equation. In each panel, from left to right, the IQR box-plots correspond to the result provided by ridge regression when the training is performed by using all SPIFs, all SPIFs except SPIF$_1$, and all SFs, respectively. The absolute errors and squared errors by varying the noise level are represented in the top and bottom panels, respectively. }    \label{fig:pulsar}
\end{figure}

PIF$_1$ in Table \ref{tab:pulsar} has a $\ast$ for the following reason. In order to study the robustness of the PIF-based approach with respect to the dimensionally-coherent combination of features, we have run the experiment twice, where, in the first case, all SPIFs were used to train the algorithm; in the second run, we have excluded SPIF$_1$ from the training phase (note that for both runs all SFs have been always included). The box-plots in Figure \ref{fig:pulsar} shows that, for both runs of the experiment, training machine learning with SPIFs led to significantly more accurate predictions than when SFs are utilized. However, the skill scores significantly worsen when SPIF$_1$, which corresponds to equation (\ref{eq:pulsar}) describing the actual physical process, is excluded from the training.

\subsection{Third example: binary system}

We then studied whether our PIF-based approach to machine learning could also be reliably applied to a classification problem. Specifically, in this third experiment our aim was to determine whether or not a pair of stars or celestial bodies were gravitationally bound, having at disposal measures of the energies of the binary system, the bound energy being defined as \citep{hilditch2001introduction}
\begin{equation}
\label{eq: grav en}
E = \frac{1}{2}\frac{m_1 \cdot m_2}{(m_1 + m_2)} \cdot v^2 - G \frac{m_1 \cdot m_2}{r},
\end{equation}
where $m_1$ and $m_2$ represent the masses of the bodies, $v$ is their relative velocity, $r$ is their distance, and $G$ is the gravitational constant. In this case, the annotation of the data archive was realized by providing label $0$ to condition $E>0$ and label $1$ to condition $E<0$. Features and PIFs considered for this experiment are in Table \ref{tab:binary-features}.

\begin{table}[h]
    \centering
    \begin{tabular}{ l c l c}
    \toprule
    {\textbf{feature}} & {\textbf{unit}} & {\textbf{PIF}} & {\textbf{unit}} \\ \midrule
    F$_1= m_1$ & $kg$ & PIF$_1=\frac{m_1 \cdot m_2}{m_1 + m_2 }\cdot v^2$ & $J$ \\ \midrule
    F$_2=m_2$ & $kg$ & PIF$_2=-G \frac{m_1 \cdot m_2}{r}$ & $J$ \\ \midrule
    F$_3=v$ & $m/s$ &  & \\ \midrule
    F$_4=r$ & $m$ & & \\ \midrule
    \end{tabular}
    \caption{Classification of binary systems. Features and PIFs with respective units.}
    \label{tab:binary-features}
\end{table}

The application of ridge-regression-based classification to the test sets of SFs and SPIFs provided the confusion matrices in equation (\ref{eq:binary-confusion}). In each one of these matrices entry $(1,1)$ contains the true positives (TPs), entry $(2,2)$ contains the true negatives (TNs), entry $(1,2)$ contains the false positives (FPs), and entry $(2,1)$ contains the false negatives (FNs). 

\begin{equation}\label{eq:binary-confusion}  
\text{CM}_{\text{SF}} = \left[ \begin{array}{cc}
1507 & 61 \\ 2 & 30
\end{array}
\right]
%\end{equation}
\ \ \ 
%\begin{equation}   
\text{CM}_{\text{SPIF}} = \left[ \begin{array}{cc}
1550 & 18 \\ 2 & 30
\end{array}
\right].
\end{equation}

\begin{table}[]
    \centering
    \begin{tabular}{r |  c  c}
    \toprule
    {\textbf{Score}} & {\textbf{SF}} & {\textbf{SPIF}} \\ \midrule
    TSS & 0.898 & 0.926 \\\midrule
    HSS & 0.472 & 0.744 \\ \midrule
    Sensitivity &0.998 & 0.998\\ \midrule
    Specificity & 0.329 & 0.625\\ \midrule
    Accuracy &0.960 & 0.987\\ \bottomrule
    \end{tabular}
    \caption{Classification of binary systems. Skill scores computed from the confusion matrices in equation (\ref{eq:binary-confusion}).}
    \label{tab:binary-classification}
\end{table}

From these confusion matrices we computed several skill scores assessing the classification effectiveness of the algorithm, i.e., the Sensitivity
\begin{equation}
{\mbox{Sensitivity}} = \frac{\mbox{TP}}{\mbox{TP}+\mbox{FN}} \ ,
\end{equation}
the Specificity
\begin{equation}
    {\text{Specificity}} = \frac{\text{TN}}{\text{TN}+\text{FP}} \ ,
\end{equation}
the accuracy
\begin{equation}  \text{Accuracy}=\frac{\text{TP}+\text{TN}}{\text{TP}+\text{TN}+\text{FP}+\text{FN}} \ ,
\end{equation}
the True Skill Statistic (TSS)
\begin{equation}
    \text{TSS}= \text{Sensitivity}+\text{Specificity}-1 \ ,
\end{equation}
and the Heidke Skill Score (HSS)
\begin{equation}
    \text{HSS} =     \frac{{2 \cdot (\text{TP} \cdot \text{TN} - \text{FN} \cdot \text{FP})}}{{(\text{TP} + \text{FN}) \cdot (\text{FN} + \text{TN}) + (\text{TP} + \text{FP}) \cdot (\text{FP} + \text{TN})}}.
\end{equation}
The values of these skill scores obtained from the entries in the confusion matrices (\ref{eq:binary-confusion}) are reported in Table \ref{tab:binary-classification}.

\section{Application to real data: solar flare forecasting}
\label{blindflare}
Solar flares are the most energetic events in the heliosphere \citep{piana2022hard}. Their significance is two-fold. On the one hand, the physical mechanisms at their base are still mostly unknown and the identification of the model equations that realistically describe them is a crucial open issue in solar physics \citep{tandberg1988physics}. On the other hand, solar flares are the main trigger of space weather, so that the realization of a reliable flare forecasting process may have a crucial impact on the safeguard of several both in space and on Earth assets \citep{camporeale2018machine,GEORGOULIS2024}. 

The use of machine learning for flare forecasting is rather recent, and typically relies on the well-established result that these explosions are generated by active regions on the solar atmosphere characterized by a complex magnetic configuration. The flare forecasting game leveraging feature-based supervised machine learning approaches is therefore typically based on the following process \citep{guastavino2023operational,guastavino2022implementation,cicogna2021flare,georgoulis2021flare,campi2019feature,florios2018forecasting}:
\begin{enumerate}
    \item The magnetograms in the database of the Helioseismic and Magnetic Imager on board the NASA Solar Dynamics Observatory \citep{schou2012design}, which represents the most complete and up-to-date archive of active regions images, are annotated by means of the light-curves measured by the GOES instrument, which reveals the flare presence by means of a continuous monitoring of the X-ray emission from the Sun.
    \item A feature extraction algorithm is applied to the magnetograms to extract large sets of active regions descriptors.
    \item A machine learning algorithm is trained by using the HMI features and the corresponding GOES binary labels.
\end{enumerate}

In this experiment we applied this process by using the nine features indicated in the first column of Table \ref{tab:flare}, which are considered among the most predictive descriptors of solar flares. These selected features represent various physical quantities, including electric current ($I$), force ($F$), magnetic helicity ($H$), magnetic flux ($\Phi$), area ($S$), energy density ($\rho$), magnetic field strength ($B$), magnetic field gradient ($\nabla B$), and characteristic length ($l$). These features have been extracted from the portion of the HMI archive in the time interval between 9/15/2012 and 09/07/2017. From these features we have generated the PIFs in the second column of the same table, where the triggering dimension is energy (we note that GOES provides also the class of the flare, which is related to the energy it releases during its explosive phase). The annotated archive made of active region magnetograms has been chronologically split into two sets made of $70\%$ and $30\%$ of the whole database, respectively. Ridge regression was trained using the first portion of the corresponding SFs and SPIFs, and validated against the second ones, to obtain the confusion matrices
\begin{equation}\label{eq:flare-confusion}  
{\text{CM}}_{\text{{SF}}} = \left[ \begin{array}{cc}
187 & 50 \\ 9 & 36
\end{array}
\right]
%\end{equation}
\ \ \ 
%\begin{equation}   
{\text{CM}}_{\text{{SPIF}}} = \left[ \begin{array}{cc}
210 & 27 \\ 11 & 34
\end{array}
\right] \ \ ,
\end{equation}
from which we computed the skill scores described in Table \ref{tab:flare-skill}. The application of the feature ranking algorithm to the PIFs in the table identified PIF$_2$ as the PIF that mostly impacts the classification process.

\begin{table}[]
    \centering
    \begin{tabular}{ l c l c }
    \toprule
        feature & unit & PIF & unit \\
        \midrule
        F$_1=I$ & $A$ & PIF$_1 = B \cdot I \cdot S $ & $TAm^2$ \\
        \midrule
      F$_2=F$ & $T\cdot A \cdot m$ & PIF$_2 =\Phi \cdot I$ & $T \cdot A \cdot m^2$       \\ 
      \midrule
      F$_3=H$ & $\frac{T^2}{m}$ & PIF$_3 =\frac{F \cdot \nabla B \cdot S}{B} $ & $T \cdot A \cdot m^2$       \\ 
      \midrule
      F$_4=\Phi$ & $T\cdot m^2$ & PIF$_4 =\frac{F\cdot B^2}{H}$ & $T \cdot A \cdot m^2$       \\ 
      \midrule
      F$_5=S$ & $m^2$ & PIF$_5 =\frac{H \cdot I \cdot S}{\nabla B}$& $T \cdot A \cdot m^2$      \\ 
      \midrule
      F$_6=\rho$ & $\frac{T \cdot A}{m}$ & PIF$_6 =\frac{H \cdot I^2 \cdot S^2}{F}$ & $T \cdot A \cdot m^2$       \\ 
      \midrule
      F$_7=B$ & $T$ & PIF$_7 =\nabla B \cdot l \cdot I \cdot S$ & $T \cdot A \cdot m^2$       \\ 
      \midrule
       F$_8=\nabla B$ & $\frac{T}{m}$ & PIF$_8 =\rho \cdot l \cdot S$ & $T \cdot A \cdot m^2$       \\ 
       \midrule
       F$_9=l$ & $m$ & &       \\ \bottomrule
    \end{tabular}
    \caption{Solar flare forecasting. Features and PIFs with respective units.}
    \label{tab:flare}
\end{table}

\begin{table}[]
    \centering
    \begin{tabular}{r | c c}
    \toprule
    Score & SF & SPIF \\
    \midrule
    TSS & 0.590 & 0.642 \\
    \midrule
    HSS & 0.430 & 0.561 \\
    \midrule
    Sensitivity & 0.954 & 0.950 \\
    \midrule
    Specificity & 0.419 & 0.557\\
    \midrule
    Accuracy & 0.791 & 0.865\\
    \midrule
    \end{tabular}
    \caption{Solar flare forecasting. Skill scores computed from confusion matrices in equation (\ref{eq:flare-confusion}).}
    \label{tab:flare-skill}
\end{table}

\section{Conclusions}
This study introduces a physics-informed feature-based approach to machine learning. In this approach, feature maps are designed to generate physics-informed features characterized by the same physical dimension. On the one hand, experiments with synthetic data showed that 1) training machine learning with these PIFs enhances both the regression and the classification performances of the algorithm; and 2) the constructed regression coefficients provide accurate estimate of the equation describing the data formation process.

An experiment on flare forecasting realized by means of real data showed that the improvement in the classification power still persists, although it is less significant. This may be explained by the fact that in flare forecasting the equations used to generate the PIFs have been designed without a full knowledge of the physics modeling the flare mechanisms, which is still largely unknown. However and rather interestingly, the ranking algorithm applied to the PIFs in this application points out PIF$_2 = \Phi I$ as the most predictive operation for flare forecasting. In fact, the product of the magnetic flux times the electric current is a measure of the so-called magnetic helicity, which is a well-investigated candidate for representing a significant portion of the energy budget stored in the active regions \citep{tziotziou2012magnetic,liokati2022magnetic,park2010productivity}. This may induce to open a new research line, where the generation of large numbers of PIFs can be utilized to possibly identify unknown descriptors of the way energy distributes in the active region as a precursor of the flaring activity.

\section*{Acknowledgments}
ML and MP acknowledge the support of the Fondazione Compagnia di San Paolo within the framework of the Artificial Intelligence Call for Proposals AIxtreme project (ID Rol: 71708). 
MP also acknowledges the financial support of the National Recovery and Resilience Plan (NRRP), Mission 4, Component 2, Investment 1.1, Call for tender No. 104 published on 2.2.2022 by the Italian Ministry of University and Research (MUR), funded by the European Union – NextGenerationEU – Project Title ``Inverse Problems in Imaging Sciences (IPIS)'' – CUP D53D23005740006 – Grant Assignment Decree No. 973 adopted on 30/06/2023 by the Italian Ministry of Ministry of University and Research (MUR). 
FB acknowledges financial support under the National Recovery and Resilience Plan (NRRP), Mission 4, Component 2, Investment 1.1, Call for tender No.1409 published on 14.9.2022 by the Italian Ministry of University and Research (MUR), funded by the European Union – NextGenerationEU– Project Title ``CORonal mass ejection, solar eNERgetic particle and flare forecaSTing from phOtospheric sigNaturEs (CORNERSTONE)'' – CUPD53D23022570001 - Grant Assignment Decree No. 1294 adopted on 4.8.2023 by the Italian Ministry of Ministry of University and Research (MUR).
All authors are also grateful to the Gruppo Nazionale per il Calcolo Scientifico - Istituto Nazionale di Alta Matematica (GNCS - INdAM).
This research has been performed in the framework of the MIUR Excellence Department Project awarded to Dipartimento di Matematica, Università di Genova, CUP D33C23001110001.

\bibliographystyle{aa}
%\bibliography{reference-list}
\bibliography{references}
\end{document}